\documentclass[letterpaper]{article} 
\usepackage[preprint]{aaai2027}  
\usepackage[hyphens]{url}  
\usepackage{graphicx} 
\usepackage{natbib}  
\usepackage{caption} 
\usepackage{algorithm}
\usepackage{algorithmic}

\usepackage{newfloat}
\usepackage{listings}
\DeclareCaptionStyle{ruled}{labelfont=normalfont,labelsep=colon,strut=off} 
\floatstyle{ruled}
\newfloat{listing}{tb}{lst}{}
\floatname{listing}{Listing}

\usepackage{booktabs}

\usepackage{amsmath}
\usepackage[table]{xcolor}
\usepackage{multirow} 
\usepackage{mathtools}
\usepackage{amssymb}
\usepackage{tabularx}
\newcommand{\yes}{\ding{51}}

\usepackage{pifont}    

  \newcommand{\gain}[1]{\textsubscript{\textcolor{gray!55!black}{\scriptsize #1}}}

\definecolor{Lavender}{RGB}{148, 198, 205}

\colorlet{tableheadcolor}{gray!10} 
\colorlet{tablerowcolor}{Lavender!10} 
\newcommand{\headcol}{\rowcolor{tableheadcolor}}
\newcommand{\rowcol}{\rowcolor{tablerowcolor}}

\title{ULME: Predictive Coupling for Mutually Reinforcing \\ Embedding and Language Modeling}
\author{
    Yang Liu\textsuperscript{\rm 1, \rm 2, \rm 3, $\dagger$}, Zhiqiang Yuan\textsuperscript{\rm 1, $\dagger$, $\S$}, Shu-Dong Huang\textsuperscript{\rm 2, \rm 3, *}, Jiancheng Lv\textsuperscript{\rm 2, \rm 3}, Liwei Yang\textsuperscript{\rm 1}, \\
    Qiwei Yan\textsuperscript{\rm 1}, Kailin Lv\textsuperscript{\rm 1}, Jinchao Zhang\textsuperscript{\rm 1, *}
}
\affiliations{
    \textsuperscript{\rm 1}Pattern Recognition Center, WeChat AI, Tencent Inc.\\
    \textsuperscript{\rm 2}College of Computer Science, Sichuan University\\
    \textsuperscript{\rm 3}Engineering Research Center of Machine Learning and Industry Intelligence, Ministry of Education\\

}

\begin{document}

\maketitle

\begingroup
\renewcommand{\thefootnote}{}
\footnotetext{
$\dagger$ Equal contribution.
$\S$ Tech Lead.
* Corresponding authors.
$\ddagger$ This work was done during Yang Liu's internship at WeChat AI, Tencent Inc. under the guidance of Zhiqiang Yuan.
}
\endgroup

\begin{abstract}
Text embedding and text generation are two of the most useful capabilities of modern large language models, yet they are typically provided by separate specialized models—an embedder and a generator—trained with incompatible objectives and deployed side by side in retrieval-augmented generation and agent systems. 
Recent unified embedding-generation models such as GritLM share parameters between the two capabilities, but they either switch attention regimes between modes or optimize embedding and language modeling as parallel objectives that never supervise each other. 
We argue that these designs are only partially unified: the two capabilities coexist on the same weights, but do not shape each other. 
We propose ULME, a post-training framework that unifies embedding and language modeling on a causal decoder-only LLM through \emph{predictive coupling}. 
ULME inserts a chain of learnable slots between the query and the answer, reads a semantic
embedding and a reasoning embedding from two designated slots, and trains  three objectives on a shared causal backbone: a contrastive loss anchoring the semantic slot to document representations, a language-modeling loss supervising answer generation, and a stop-gradient predictive loss that aligns the reasoning slot with a pooled summary of the answer, so that generation actively shapes what the embedding encodes. 
Across two model families (Qwen2.5, Qwen3) and eight scales spanning 0.5B–8B parameters, a single ULME checkpoint improves MTEB by 3.3–4.1 points over the strongest open-source baseline, tightens BRIGHT and AdvBench-IR simultaneously, and remains on par with a generation-only fine-tune on standard accuracy benchmarks (MMLU, GSM8K, MMLU-Pro, HumanEval)—demonstrating that genuine unification yields mutually reinforcing rather than merely coexisting capabilities.
\end{abstract}


\section{Introduction}

Modern LLM systems routinely process the same query twice.
In retrieval-augmented generation~\cite{lewis2020retrieval,yan2026rava}, in-context example selection~\cite{liu2022makes}, and agent
memory~\cite{xu2026mem,lyu2026photocraft}, an embedder first maps the query into a vector to retrieve context, and a generator then processes the query again
to produce an answer.
The two components are typically trained separately, and their representations of the same query never interact.
This treats the semantic understanding needed to retrieve as if it were independent of the semantic understanding needed to generate---yet a
generative LLM, on the way to producing a coherent answer, already encodes the query in a form that is largely what a good embedding should
be.

More precisely, generating an appropriate answer requires the model to encode, in its causal hidden states, an answer-relevant reading of the
query.
This reading is not yet an embedding: it is distributed across positions, has no fixed-dimensional readout, and is not organized into a
metric space in which similarity reflects relevance.
Recent LLM-based embedders~\cite{zhang2025qwen3,lee2025nv} confirm the underlying assumption: competitive dense representations can be
extracted directly from a decoder-only LLM's causal hidden states, without a separate encoder or bidirectional attention.

Exposing these states as embeddings is not as simple as applying both losses to the same token, however.
Retrieval and language modeling inhabit the same hidden space but impose different geometries on it: retrieval asks \emph{where an input
should lie relative to others}, requiring a sequence-level state that is globally comparable under inner product; language modeling asks
\emph{which token should come next}, requiring position-dependent states supporting the LM head.
Without a dedicated readout, the natural embedding candidate is the final hidden state of the query---yet that state is also used to predict
the first answer token.
Overloading it creates a \emph{representational collision}: one surface-token state is simultaneously asked to be a globally comparable
embedding and a local next-token predictor.
A unified model should separate the two readouts, but couple the semantics they share.

Recent unified models---GritLM~\cite{muennighoff2025generative}, GEM~\cite{zhang2025gem}---achieve parameter-level unification but leave two
gaps.
First, the two capabilities may form their representations under different computations: GritLM, for example, switches to bidirectional
attention for embedding, so its embedding readout and its generative hidden states do not belong to the same causal trajectory.
Second, the two objectives are optimized in parallel: generation does not depend on the learned embedding, and embedding receives no direct
signal from the answer-side representations produced by generation.
Parameter sharing alone does not establish an interface between them.

We close both gaps by treating the causal hidden-state trajectory as the shared latent, and by introducing a mechanism we call
\emph{predictive coupling}: the answer-side hidden states from the same forward pass---a future-informed view of the query---serve as a
learning target for a designated query-side latent, aligned by an InfoNCE loss with a stop-gradient on the answer side.
The stop-gradient makes the coupling well-posed: the query-side latent reaches toward the more-informed answer-side representation, while
gradient never flows back into the answer side that language modeling separately supervises.
This is a one-sided distillation from the answer-side representation into a query-only latent that is available at inference time.
Realizing it on the causal trajectory requires dedicated latent positions---slots that summarize the query without being tied to any surface
word---so both the retrieval readout and the predictive target can be trained without competing with next-token prediction.

Building on these ideas, we propose \textbf{ULME} (\textbf{U}nifying \textbf{L}anguage \textbf{M}odeling and \textbf{E}mbedding), a
post-training framework in which a decoder-only LLM, under causal attention throughout, produces embedding and answer from a single forward
pass over $q \,\Vert\, \langle\textsc{emb}\rangle_{1:N} \,\Vert\, a$.
The $N$ learnable slot tokens realize the shared latent: slot $0$ is the \emph{semantic embedding}, aligned to document representations by
$\mathcal{L}_{\text{sem}}$; the answer span is supervised by $\mathcal{L}_{\text{gen}}$, whose gradient reaches the slots via the answer's
attention over them; slot $N{-}1$ is the \emph{reasoning embedding}, aligned by a stop-gradient predictive loss $\mathcal{L}_{\text{pred}}$
to a pooled summary of the answer-side hidden states.
All three losses share every parameter and every hidden state; embedding is no longer trained in isolation, but shaped by the representation
the model would use to answer the query.

Our contributions are:
\begin{itemize}
\item We reframe unified embedding and generation as reconciling two geometries within one causal hidden space---a globally comparable
sequence-level state for retrieval, and position-dependent states for next-token prediction---motivating separate readouts with coupled
semantics.
\item We propose ULME with \emph{predictive coupling}: under stop-gradient, an answer-side latent from the same forward pass supervises a
query-side latent, so generation shapes embedding without being disturbed by it. A learnable slot chain between query and answer hosts the
semantic readout at one end and the predictive target at the other.
\item Across two model families (Qwen2.5, Qwen3) and eight scales (0.5B--8B), a single ULME checkpoint is simultaneously competitive on
embedding (MTEB, BRIGHT) and generation (MMLU, GSM8K, MMLU-Pro, HumanEval), improving over the strongest same-backbone baseline on MTEB by
3.3--4.1 points.
\end{itemize}

\begin{figure*}[!t]
\begin{center}
\includegraphics[width=\linewidth]{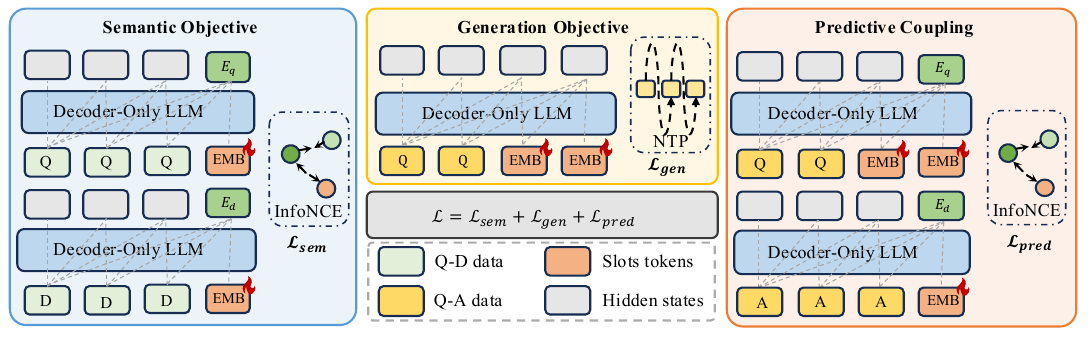}
\end{center}
\caption{Overview of ULME. A single causal decoder is trained with three objectives on a shared chain of learnable slot tokens (flame icons: trainable slots; grey blocks: hidden states).
\textbf{Left} ($\mathcal{L}_{\text{sem}}$): symmetric InfoNCE pulls the query and document slot embeddings $E_q, E_d$ together.
\textbf{Middle} ($\mathcal{L}_{\text{gen}}$): next-token prediction over the answer span $A$, with the slot chain on the causal path from $Q$ to $A$.
\textbf{Right} ($\mathcal{L}_{\text{pred}}$): a one-sided InfoNCE aligns the query-side slot with the answer-side slot under stop-gradient on the answer side. Total loss: $\mathcal{L} = \mathcal{L}_{\text{sem}} + \mathcal{L}_{\text{gen}} + \mathcal{L}_{\text{pred}}$.}
\label{pipline}
\vspace{-0.5em}
\end{figure*}

\section{Related Work}
\paragraph{Large Language Models.} 
Recent years have witnessed remarkable progress in large language models (LLMs), which have demonstrated unprecedented capabilities in understanding and generating human-like text and have come to dominate a wide range of natural language processing tasks. 
State-of-the-art LLMs such as GPT \cite{achiam2023gpt}, 
Llama \cite{grattafiori2024llama}, 
DeepSeek \cite{xu2026deepseek}, 
and Qwen \cite{yang2025qwen3} are all built on the decoder-only transformer architecture. 
The decoder-only design relies on causal attention, in which each token attends only to its predecessors, and is trained with a next-token prediction objective~\cite{radford2018improving}; this combination has proven crucial to the scalability and downstream versatility of modern LLMs. 
Beyond their core generative ability, these models exhibit strong instruction following~\cite{ouyang2022training}, few-shot in-context learning~\cite{brown2020language}, and multi-step reasoning~\cite{wei2022chain}, and have been extended in two complementary directions: retrieval-augmented generation~\cite{lewis2020retrieval}, which couples an LLM with an external knowledge base to mitigate knowledge cutoff and reduce hallucination; and LLM agents~\cite{yao2022react}, which equip LLMs with tools, memory, and environment interaction to perform long-horizon decision making. 
Both of these downstream paradigms rely on a retriever that selects relevant context for the generator, yet the retriever and the generator are typically separate models trained with separate objectives — a separation that motivates the unified embedding-generation setting we study in this work.

\paragraph{LLM-based Embedding Models.} 
Text embedding models map text into dense vectors and underpin retrieval, clustering, and classification~\cite{reimers2019sentence,neelakantan2022text}. 
Early models were built on bidirectional encoders such as BERT~\cite{devlin2019bert}, but as decoder-only LLMs surpassed them in language understanding~\cite{brown2020language,touvron2023llama}, a growing line of work repurposes LLMs as embedders to inherit their pre-trained semantic knowledge~\cite{wang2022text,lee2025nv}.
E5-Mistral~\cite{wang2022text} and NV-Embed~\cite{lee2025nv} fine-tune causal LLMs with contrastive objectives and pool a single hidden state — typically the last token — as the embedding, showing that causal attention is no barrier to strong dense representations. 
Echo-Embeddings~\cite{springer2025repetition} repeats the input so earlier tokens can encode later context, while LLM2Vec~\cite{behnamghader2024llm2vec, behnamghader2026llm2vec} removes the causal mask altogether to obtain a bidirectional encoder before contrastive fine-tuning.

These methods establish the observation our work builds on: the hidden states of a decoder-only LLM, produced during an ordinary causal forward pass, already carry enough information to serve as strong embeddings. Yet they treat embedding as the sole objective; after contrastive fine-tuning the resulting models are competitive embedders but no longer usable as generators, giving up the generation-side capacity of the backbone rather than exploiting it.

\paragraph{Generative Retrieval.} 
Retrieval itself is increasingly posed as a generation problem rather than a pure vector-matching one. F2RVLM~\cite{bi2026f2rvlm} trains a vision-language model to \emph{generate} the fragment of a long multi-modal dialogue that answers a retrieval query, showing that a generative model can also act as the retriever when relevance is spread over multiple utterances and images; its extended study introduces the fine-grained fragment retrieval task, where a two-stage system further indexes semantic fragments with a dedicated fragment embedding model before generative retrieval~\cite{bi2026ffr}. Practical retrieval stacks thus keep an embedder and a generator side by side---one supplies a metric space for indexing, the other reasons about what the query actually asks for---which is exactly the duplication a unified embedding-generation model should absorb.

\paragraph{Unified Embedding-Generation Models.} 
Since generative LLMs and LLM-based embedders evolve in opposite directions — each discarding what the other keeps — a natural question is whether one model can retain both. 
GritLM~\cite{muennighoff2025generative} fine-tunes a single LLM with both a contrastive embedding loss and a generative language-modeling loss, switching to bidirectional attention for embedding and back to causal attention for generation. 
GEM~\cite{zhang2025gem} injects embedding ability into a generative LLM through additional tokens and a self-supervised contrastive scheme. These models show that the two capabilities can share one backbone without severe degradation on either side.

The unification they achieve, however, is partial along two axes. 
At the mechanism level, embedding and generation do not share a common forward computation, as GritLM alternates between two attention regimes. 
At the objective level, the two losses are optimized in parallel over shared parameters but never supervise each other, so the embedding remains a conventional semantic-matching vector and the generative pathway is never used to inform what it encodes. 
Our work departs from this paradigm by unifying both tasks under a single causal computation and coupling their objectives through a shared latent state, so that generation actively shapes the embedding rather than merely coexisting with it.

\section{Unified Language Modeling and Embedding}

\subsection{Model Architecture}
\label{sec:arch}

ULME operates as a single decoder-only LLM under causal attention: both the query embedding and the next-token logits are read from one forward pass over shared parameters, without mode switching or an auxiliary encoder. 
This choice fixes the mechanism axis, but it leaves open
the question of where the embedding is read from.

The standard practice in causal LLM embedders is last-token pooling. 
Under unification this is awkward: the final position of $q$ is precisely where $\mathcal{L}_{\text{gen}}$ predicts the first token of $a$, forcing a single hidden state to serve simultaneously as a semantic summary of $q$ and as a next-token predictor. 
We resolve this by inserting a chain of dedicated latent positions between $q$ and $a$, in the spirit of latent chain-of-thought approaches such as Coconut~\citep{hao2024training} and pause tokens~\citep{goyal2024think}.
These works show that decoder-only models benefit from intermediate positions where computation is not tied to any surface token; here, we
repurpose the same idea to expose readouts that are structurally addressable by every loss.

Concretely, we append $N$ learnable slot tokens $\langle\textsc{emb}\rangle_1, \ldots, \langle\textsc{emb}\rangle_N$ after $q$. A training
example takes the form
\begin{equation}
q \;\Vert\; \langle\textsc{emb}\rangle_1 \cdots \langle\textsc{emb}\rangle_N \;\Vert\; a,
\label{eq:seq}
\end{equation}
and a document is encoded analogously as $d \,\Vert\, \langle\textsc{emb}\rangle$. The slots are ordinary positions in the causal forward
pass: each $\langle\textsc{emb}\rangle_i$ attends backward to $q$ and to earlier slots, and every answer token attends to all slots. No
masking rule is altered and no auxiliary branch is introduced.

Every loss below is computed from a single causal forward pass. Let $h_q^{(i)}$ denote the hidden state at slot $i$ on the query side, $h_d$ the hidden state at $d$'s slot, and $H_a$ the answer-span hidden states. Slot $0$ carries the semantic embedding, slot $N{-}1$ carries the reasoning embedding, and $H_a$ acts as the generation target and as a teacher for slot $N{-}1$.

\subsection{Retrieval and Generation Objectives}
\label{sec:sem-gen}

We first equip the slot chain with the two objectives that any unified embedding--generation model would maintain: a contrastive loss on the
semantic slot, and a language-modeling loss on the answer.

\paragraph{Semantic retrieval.}
We designate slot $0$ as the semantic embedding: $h_q \coloneqq h_q^{(0)}$ on the query side and $h_d$ on the document side. With temperature
$\tau$ and cross-rank in-batch negatives, the semantic loss is the symmetric InfoNCE
  \begin{equation}
  \begin{aligned}
  \mathcal{L}_{\text{sem}} = &-\tfrac{1}{2B}\sum_{i=1}^{B}\log\frac{\exp(h_{q_i}^\top h_{d_i} / \tau)}{\sum_{j=1}^{B} \exp(h_{q_i}^\top
  h_{d_j}/\tau)} \\
  &-\tfrac{1}{2B}\sum_{i=1}^{B}\log\frac{\exp(h_{d_i}^\top h_{q_i} / \tau)}{\sum_{j=1}^{B} \exp(h_{d_i}^\top h_{q_j}/\tau)}.
  \end{aligned}
  \label{eq:lsem}
  \end{equation}

\paragraph{Answer generation.}
The generative loss is standard next-token cross-entropy over the full sequence, with the loss masked to the span of $a$:
\begin{equation}
\mathcal{L}_{\text{gen}} = -\sum_{t \in a} \log p_\theta\!\left(a_t \mid q,\, \langle\textsc{emb}\rangle_{1:N},\, a_{<t}\right).
\label{eq:lgen}
\end{equation}
Slot positions do not contribute to the token-level cross-entropy. Because the slot chain lies on the same causal trajectory that produces
$a$, the language-modeling gradient reaches slot hidden states through the answer's attention over them.

These two losses supervise their respective readouts under shared parameters, mirroring the parallel-objective design used by prior unified
models. What they do not provide is a direct signal that shapes the geometry of the slot chain itself: $\mathcal{L}_{\text{sem}}$ pins slot
$0$ against the document embedding but says nothing about the answer content, and $\mathcal{L}_{\text{gen}}$ conditions on the slot chain but
never explicitly asks any slot to represent anything in particular. 

\begin{table*}[t]
\centering
\small
\setlength{\tabcolsep}{5pt}
\begin{tabular}{l c c c c c c c c}
\toprule
\headcol \textbf{Method} &
\textbf{Retr.} & \textbf{Rerank.} & \textbf{Clust.} &
\textbf{PairCls.} & \textbf{Class.} & \textbf{STS} & \textbf{Summ.} & \textbf{Avg.} \\
\midrule
\multicolumn{9}{c}{\textbf{Qwen3-1.7B}} \\
\midrule
Echo~\cite{springer2025repetition}         &  4.4 & 37.7 & 38.1 & 59.3 & 70.4 & 52.1 & -0.8 & 39.8 \\
HyDE~\cite{gao2023precise}                 & 17.7 & 39.5 & 38.2 & 69.4 & 75.6 & 72.2 & 18.0 & 49.8 \\
InBedder~\cite{peng2024answer}             & 15.7 & 41.2 & 47.0 & 68.6 & \textbf{79.2} &  65.9 &  9.3 & 50.2 \\
GIRCSE~\cite{tsai2025let}                  & 27.3 & 40.4 & 47.2 & 78.2 & 70.6 & 66.6 & 22.2 & 52.5 \\
LLM2Vec~\cite{behnamghader2024llm2vec}     & 34.9 & 39.9 & 41.1 & 76.4 & 71.1 & 73.4 & 30.4 & 54.8 \\
LLM2VEC-GEN~\cite{behnamghader2026llm2vec} & 38.3 & 44.1 & \textbf{49.9} & 74.8 & 74.7 & 76.1 & 26.0 & 58.6 \\
\rowcol ULME (ours)                                & \textbf{46.5}\gain{+8.2} & \textbf{45.6}\gain{+1.5} & 48.4\gain{-1.5} & \textbf{80.1}\gain{+5.3}
& 74.1\gain{-0.6} & \textbf{78.5}\gain{+2.4} & \textbf{33.0}\gain{+7.0} & \textbf{61.9}\gain{+3.3} \\
\midrule
\multicolumn{9}{c}{\textbf{Qwen3-4B}} \\
\midrule
Echo~\cite{springer2025repetition}         &  7.6 & 38.7 & 39.7 & 64.1 & 73.5 & 58.7 &  7.6 & 43.6 \\
HyDE~\cite{gao2023precise}                 & 14.0 & 36.9 & 36.4 & 47.5 & \textbf{79.9} & 57.9 & 26.7 & 44.7 \\
InBedder~\cite{peng2024answer}             & 13.4 & 41.8 & 48.1 & 68.8 & 79.6 & 64.9 & 26.1 & 50.1 \\
GIRCSE~\cite{tsai2025let}                  & 27.8 & 42.9 & 45.1 & 70.7 & 71.5 & 63.6 & 22.5 & 51.3 \\
LLM2Vec~\cite{behnamghader2024llm2vec}     & 41.1 & 40.0 & 43.1 & 78.5 & 72.5 & 71.6 & 31.1 & 56.8 \\
LLM2VEC-GEN~\cite{behnamghader2026llm2vec} & 39.8 & 45.9 & \textbf{50.6} & 78.1 & 77.2 & 78.6 & 28.7 & 60.6 \\
\rowcol ULME (ours)                                & \textbf{48.1}\gain{+8.3} & \textbf{47.4}\gain{+1.5} & 49.1\gain{-1.5} & \textbf{83.6}\gain{+5.5}
& 76.6\gain{-0.6} & \textbf{81.1}\gain{+2.5} & \textbf{36.4}\gain{+7.7} & \textbf{64.1}\gain{+3.5} \\
\midrule
\multicolumn{9}{c}{\textbf{Qwen3-8B}} \\
\midrule
Echo~\cite{springer2025repetition}         &  6.8 & 40.0 & 37.2 & 63.6 & 74.2 & 53.9 &  0.5 & 41.8 \\
HyDE~\cite{gao2023precise}                 & 15.9 & 37.1 & 32.4 & 65.8 & \textbf{81.6} & 67.4 & 30.3 & 48.3 \\
InBedder~\cite{peng2024answer}             & 11.0 & 42.4 & 48.6 & 70.7 & 80.5 & 67.4 & 24.5 & 50.5 \\
GIRCSE~\cite{tsai2025let}                  & 36.3 & 41.3 & \textbf{50.9} & 74.7 & 74.2 & 68.8 & 26.6 & 56.5 \\
LLM2Vec~\cite{behnamghader2024llm2vec}     & 42.7 & 40.9 & 40.6 & 77.3 & 72.5 & 72.6 & 31.7 & 56.8 \\
LLM2VEC-GEN~\cite{behnamghader2026llm2vec} & 43.4 & 46.4 & 49.8 & 80.6 & 77.6 & 79.7 & 32.1 & 61.9 \\
 \rowcol ULME (ours)                                & \textbf{52.3}\gain{+8.9} & \textbf{48.2}\gain{+1.8} & 50.0\gain{+0.2} & \textbf{86.3}\gain{+5.7}
& 77.1\gain{-0.5} & \textbf{82.2}\gain{+2.5} & \textbf{40.7}\gain{+8.6} & \textbf{66.0}\gain{+4.1} \\
\bottomrule
\end{tabular}
\caption{Results on MTEB(eng, v2). Rows are grouped by backbone; columns 2--8 are per-category means over Retrieval, Reranking, Clustering,
PairClassification, Classification, STS, and Summarization; the final column is the overall task-mean over all 41 tasks. Bold marks the best
score in each column within a backbone group.}
\label{tab:mteb}
\end{table*}

\subsection{Predictive Coupling}
\label{sec:pred}

$\mathcal{L}_{\text{gen}}$ is a token-level, indirect signal on the query side: it enforces that $a$ can be produced from $\{q,
\langle\textsc{emb}\rangle_{1:N}\}$, but it never demands that any specific slot representation take a particular form. To place a direct,
sequence-level constraint on the query side, we align the terminal slot with a pooled summary of the answer computed in the same forward
pass.

Let $r_{qa}$ denote a pooled latent obtained from the answer-side hidden states $H_a$; we treat the pooling as a fixed operator and defer its
exact form to the Experiments section (throughout this paper $r_{qa}$ is the mean of $H_a$; alternatives are ablated). The predictive loss is a one-sided in-batch InfoNCE between $h_q^{(N-1)}$ and $r_{qa}$:
\begin{equation}
\mathcal{L}_{\text{pred}} = -\log \frac{\exp\!\big(\langle h_q^{(N-1)},\, \operatorname{sg}(r_{qa})\rangle / \tau\big)}{\sum_k
\exp\!\big(\langle h_q^{(N-1)},\, \operatorname{sg}(r_{qa,k})\rangle / \tau\big)},
\label{eq:lpred}
\end{equation}
where $\{r_{qa,k}\}$ enumerates the (cross-rank) batch and $\operatorname{sg}(\cdot)$ is the stop-gradient operator.

The stop-gradient placement is deliberate. $r_{qa}$ has causal access to both $q$ and $a$ and is strictly more informed than $h_q^{(N-1)}$,
which sees only $q$ and the slot chain; a well-posed predictive objective should let the less-informed side reach toward the more-informed
one, not the reverse. Allowing gradients to flow back into $r_{qa}$ would additionally push the answer-side representation toward whatever is
easy to contrast in a batch, distorting the hidden states that $\mathcal{L}_{\text{gen}}$ must simultaneously supervise for next-token
prediction.

$\mathcal{L}_{\text{pred}}$ is complementary to $\mathcal{L}_{\text{gen}}$ rather than redundant with it. The generative loss shapes slot
representations indirectly, at token granularity, through the causal dependencies of $a$ on the chain. The predictive loss shapes slot
$N{-}1$ directly, at sequence granularity, by requiring it to be discriminative for the pooled answer. Together, the two losses populate the
slot chain from both sides --- one from the token stream of $a$, one from the compressed summary of $a$ --- while $\mathcal{L}_{\text{sem}}$
anchors slot $0$ to the document space.

The full ULME objective is
\begin{equation}
\mathcal{L} = \lambda_{\text{sem}}\,\mathcal{L}_{\text{sem}} + \lambda_{\text{gen}}\,\mathcal{L}_{\text{gen}} +
\lambda_{\text{pred}}\,\mathcal{L}_{\text{pred}},
\label{eq:total}
\end{equation}
computed from a single causal forward pass over $q \,\Vert\, \langle\textsc{emb}\rangle_{1:N} \,\Vert\, a$ (and $d \,\Vert\,
\langle\textsc{emb}\rangle$ on the document side).

\section{Experiments}
\label{sec:experiments}
Our experiments aim to answer one central question: does the joint objective of ULME actually deliver both capabilities at once, or merely time-share them? 
To make the answer decisive, every reported number in this section --- embedding \emph{and} generation --- comes from a single trained ULME checkpoint per configuration; no head is fine-tuned in isolation, and no capability is toggled off at evaluation time.

We instantiate ULME on two open-weight decoder-only LLM families---Qwen2.5~\cite{team2024qwen2} and Qwen3~\cite{zhang2025qwen3}---at eight scales spanning 0.5B to 8B (Qwen2.5-\{0.5B, 1.5B, 3B, 7B\}-Instruct and Qwen3-\{0.6B, 1.7B, 4B, 8B\}), and evaluate each trained model along both axes. For space reasons the main tables in this section report only three representative Qwen3 scales (1.7B, 4B, 8B); the remaining five backbones follow the same trends and their full results are provided in the \emph{supplementary material}. On the embedding side we use MTEB~\cite{muennighoff2023mteb},
the standard general-purpose text embedding benchmark, and BRIGHT~\cite{su2025bright},
a reasoning-intensive retrieval benchmark whose queries require inference beyond surface matching. On the generation side we use MMLU~\cite{hendrycks2020measuring}
for broad knowledge, GSM8K~\cite{cobbe2021training}
for arithmetic reasoning, MMLU-Pro~\cite{wang2024mmlu}
for harder multi-step reasoning, and
HumanEval~\cite{chen2021evaluating}
(with its stricter HumanEval+ extension~\cite{liu2023your})
for code synthesis.
This coverage stresses each half of the model with a distinct workload profile --- dense-representation quality on one side, chain-of-thought and code generation on the other --- so that a competitive score on either axis cannot be borrowed from the other.

\paragraph{Implementation details.}
Unless otherwise noted, every ULME model uses $N{=}4$ embedding slots, temperature $\tau{=}0.05$, and loss weights $\lambda_{\text{sem}}{=}\lambda_{\text{gen}}{=}1.0$, $\lambda_{\text{pred}}{=}0.3$; $r_{qa}$ is the mean of the answer-side hidden states. We train on the LLM2Vec-format E5 contrastive data~\cite{wang2022text, behnamghader2024llm2vec} for the embedding side and OpenThoughts-114k~\cite{guha2025openthoughts} for the generation side, using AdamW at $1{\times}10^{-5}$ with cross-rank in-batch negatives. Every reported number comes from the final checkpoint of a single run, without any post-hoc capability toggle. Full hyperparameters, evaluation protocols, and per-benchmark details are in the \emph{supplementary material}.

\begin{table*}[t]
\centering
\small
\setlength{\tabcolsep}{15pt}
\begin{tabular}{l c c c c c}
\toprule
\headcol \textbf{Model} &
\textbf{MMLU} & \textbf{GSM8K} & \textbf{MMLU-Pro} &
\textbf{HumanEval} & \textbf{HumanEval+}\\
\midrule
\multicolumn{6}{c}{\textbf{Qwen3-1.7B}} \\
\midrule
\headcol Backbone    & 63.7 & 76.0 & 39.3 & 54.9 & 48.2 \\
Gen-only FT                    & 63.0                       & 74.2                       & 38.6                       & 53.7
   & 46.3 \\
\rowcol ULME (ours)                    & \textbf{63.5}\gain{+0.5} & \textbf{75.3}\gain{+1.1} & \textbf{39.6}\gain{+1.0} & \textbf{56.1}\gain{+2.4} &
\textbf{49.4}\gain{+3.1} \\
\midrule
\multicolumn{6}{c}{\textbf{Qwen3-4B}} \\
\midrule
\headcol Backbone    & 72.8 & 87.9 & 52.9 & 68.3 & 62.2 \\
Gen-only FT                    & 72.2                       & 82.4                       & 51.3                       & 66.5
   & 61.0 \\
\rowcol ULME (ours)                    & \textbf{72.6}\gain{+0.5} & \textbf{86.3}\gain{+3.8} & \textbf{53.3}\gain{+2.0} & \textbf{69.5}\gain{+3.0} &
\textbf{64.0}\gain{+3.0} \\
\midrule
\multicolumn{6}{c}{\textbf{Qwen3-8B}} \\
\midrule
\headcol Backbone    & 74.8 & 91.2 & 61.8 & 71.3 & 66.5 \\
Gen-only FT                    & \textbf{75.1}              & 78.6                       & 60.2                       & 67.7
   & 62.8 \\
\rowcol ULME (ours)                    & 75.1\gain{-0.1}          & \textbf{89.5}\gain{+10.8} & \textbf{61.9}\gain{+1.7} & \textbf{72.6}\gain{+4.9} &
\textbf{67.7}\gain{+4.9} \\
\bottomrule
\end{tabular}
\caption{Generation results. \textbf{Backbone} (greyed) is the pre-training checkpoint we start from: Qwen2.5-Instruct~\cite{team2024qwen2}
or Qwen3 non-thinking~\cite{yang2025qwen3}, not trained by us. \textbf{Gen-only FT} fine-tunes the same backbone on the same
OpenThoughts-114k data used by ULME. Bold marks the better of Gen-only FT and ULME per column.}
\label{tab:gen}
\end{table*}

\begin{table}[h]
\centering
\small
\setlength{\tabcolsep}{5pt}
\begin{tabular}{l l c c}
\toprule
\headcol \textbf{Backbone} & \textbf{Method} &
\textbf{AdvBench-IR} $\downarrow$ & \textbf{BRIGHT} $\uparrow$ \\
\midrule
\multirow{3}{*}{Qwen3-1.7B}
& LLM2Vec      & 46.7           & 14.0 \\
& LLM2VEC-GEN  & 36.2  & 15.6  \\
\rowcol & ULME (ours)  & \textbf{34.9}\gain{-1.3} & \textbf{16.3}\gain{+0.7}  \\
\midrule
\multirow{3}{*}{Qwen3-4B}
& LLM2Vec      & 50.8           & 15.7 \\
& LLM2VEC-GEN  & 42.5 & 18.8  \\
\rowcol & ULME (ours)  & \textbf{40.9}\gain{-1.6}  & \textbf{19.6}\gain{+0.8} \\
\midrule
\multirow{3}{*}{Qwen3-8B}
& LLM2Vec      & 54.2           & 14.9 \\
& LLM2VEC-GEN  & 45.0  & 20.2  \\
\rowcol & ULME (ours)  & \textbf{43.2}\gain{-1.8}  & \textbf{21.4}\gain{+1.2}  \\
\bottomrule
\end{tabular}
\caption{Reasoning-intensive retrieval on BRIGHT (nDCG@10, higher is better) and AdvBench-IR (lower is better). Bold marks the best result in each column within a backbone group.}
\label{tab:bright}
\end{table}

\subsection{Main Results}
\label{sec:main}

We ask whether ULME, within a \emph{single} trained checkpoint, can match dedicated embedders on retrieval-oriented benchmarks while remaining on par with a generation-only fine-tune of the same backbone on standard accuracy benchmarks (MMLU, GSM8K, MMLU-Pro, HumanEval).
Every row of every table in this subsection is populated from the same set of weights per configuration: no capability is toggled off, no head is fine-tuned in isolation, and no mode flag is switched at evaluation time. We compare ULME against recent embedding models on the \emph{same} backbone. For LLM2Vec, LLM2VEC-GEN, and the other embedding baselines, we use the numbers reported by LLM2VEC-GEN~\cite{behnamghader2026llm2vec} on the matching backbone; backbones for which LLM2VEC-GEN does not report a given baseline are omitted from that comparison. Any observed gap is thus attributable to the training objective rather than to backbone strength or independent reproduction.

\paragraph{MTEB.}
Table~\ref{tab:mteb} reports MTEB(eng, v2) scores at three representative Qwen3 scales (1.7B, 4B, 8B); the remaining five backbones (Qwen2.5-\{0.5B, 1.5B, 3B, 7B\} and Qwen3-0.6B) follow the same trends and appear in the \emph{supplementary material}. ULME beats every baseline we train on the same backbone, reaching \textbf{66.0} on Qwen3-8B---an improvement of \textbf{9.2} points over LLM2Vec and \textbf{4.1} points over LLM2VEC-GEN on the same backbone. Against LLM2Vec, the gain persists at every scale: \textbf{+7.1 / +7.3 / +9.2} points on Qwen3-\{1.7B, 4B, 8B\}; against the stronger LLM2VEC-GEN, \textbf{+3.3 / +3.5 / +4.1}. The lifts are category-wide: on Qwen3-8B, Retrieval improves by \textbf{+9.6} over LLM2Vec, PairClassification by \textbf{+9.0}, Summarization by \textbf{+9.0}, STS by \textbf{+9.6}, and Clustering by \textbf{+9.4}, with the remaining categories in the \textbf{+4--7} range. The largest lifts (Retrieval, Summarization) are on tasks whose target is itself a produced text---exactly the setting where supervising the query latent through an answer-side representation should help.

\paragraph{BRIGHT and AdvBench-IR.}
Table~\ref{tab:bright} probes the reasoning-heavy side. On BRIGHT (nDCG@10, higher is better) ULME improves over LLM2VEC-GEN by \textbf{+0.7 / +0.8 / +1.2} points on Qwen3-\{1.7B, 4B, 8B\}, with the gap widening as scale grows---a stronger backbone extracts more from the coupling, and the predictive objective is what converts that extra capacity into inference-heavy retrieval accuracy. AdvBench-IR (lower is better) moves in the same direction: ULME cuts it by \textbf{25.3\%} on Qwen3-1.7B and \textbf{20.3\%} on Qwen3-8B relative to LLM2Vec, tightening LLM2VEC-GEN's own reduction by another \textbf{1.3--1.8} points at every scale.Both axes moving in ULME's favor rules out a one-property-for-another trade-off.

\paragraph{Generation.}
Table~\ref{tab:gen} reports the same ULME checkpoints on MMLU, GSM8K, MMLU-Pro, and HumanEval / HumanEval\texttt{+}, alongside the untrained backbone and a generation-only fine-tune on the same OpenThoughts-114k data used to train ULME. Since Backbone is an external reference, the relevant comparison for whether joint training holds up on these accuracy benchmarks is against Gen-only FT. On knowledge and math (MMLU, GSM8K, MMLU-Pro), ULME matches or exceeds the fine-tune on \textbf{all cells but one} (an $-0.1$ tie on Qwen3-8B MMLU), with MMLU-Pro gains that grow with scale (\textbf{+1.0 / +2.0 / +1.7} at 1.7B / 4B / 8B) and GSM8K gains that reach \textbf{+10.8} at 8B. On code, ULME is above the fine-tune at every scale, by \textbf{+2.4 to +4.9} points on HumanEval and \textbf{+3.0 to +4.9} on HumanEval\texttt{+}, largest at Qwen3-8B (HE\texttt{+}: 67.7 vs 62.8). Consistent with the role of $\mathcal{L}_{\text{pred}}$, forcing the query slot to summarize what the model would generate is especially informative for structured, functionally-verified outputs where the pooled answer summary discriminates between many otherwise plausible completions. Both ULME and the fine-tune occasionally trail Backbone, which we attribute to distributional shift from continued training on OpenThoughts-114k rather than to the joint objective---the same pattern appears in Gen-only FT.

  \begin{table}[t]
  \centering
  \small
  \setlength{\tabcolsep}{1.3mm}

  \begin{tabular}{l c c c}
  \toprule
  \headcol \textbf{Slots} & \textbf{MTEB-lite} &
  \textbf{MMLU} & \textbf{GSM8K} \\
  \midrule
  $N=0$ & 58.6 & 74.5 & 87.8 \\
  $N=1$            & 62.4 & 76.1 & 89.4 \\
  $N=2$            & 63.1 & 76.4 & 89.8 \\
  \rowcol $N=4$   & \textbf{63.8} & \textbf{76.5} & \textbf{90.0} \\
  $N=8$            & 63.6 & 76.3 & 89.9 \\
  \bottomrule
  \end{tabular}
  \caption{Ablation on the number of embedding slots $N$. Highlighted row is our default setting.}
  \label{tab:abl_slot}
  \end{table}

\begin{table}[t]
  \centering
  \small
  \setlength{\tabcolsep}{0.5mm}
  \begin{tabular}{c c c c c c }
  \toprule
  \headcol $\mathcal{L}_{\text{sem}}$ & $\mathcal{L}_{\text{gen}}$ & $\mathcal{L}_{\text{pred}}$
  & \textbf{MTEB-lite}
  & \textbf{MMLU} & \textbf{GSM8K} \\
  \midrule
   & \yes & \yes & 31.2 & \textbf{76.6} & \textbf{90.1}  \\
  \yes &  & \yes & 61.0 & 61.8 & 74.2 \\
  \yes & \yes &  & 60.2 & 76.4 & 89.9  \\
  \rowcol \yes & \yes & \yes & \textbf{63.8} & 76.5 & 90.0 \\

  \bottomrule
  \end{tabular}
  \caption{Loss decomposition. \yes\ = loss enabled, blank = removed. Highlighted row is our full model.}
  \label{tab:abl_loss}
  \vspace{-0.5em}
  \end{table}
  
  \begin{table}[t]
  \centering
  \small
  \setlength{\tabcolsep}{5pt}
    \begin{tabular}{l c c c}
  \toprule
  \headcol $r_{qa}$ \textbf{pooling} & \textbf{MTEB-lite} &
  \textbf{MMLU} & \textbf{GSM8K}  \\
  \midrule
  \rowcol Mean over $H_a$ & \textbf{63.8} & \textbf{76.5} & \textbf{90.0}   \\
  Last-token of $a$         & 61.9 & 76.3 & 89.8  \\
  Length-weighted mean      & 62.7 & 76.4 & 89.9  \\
  \bottomrule
  \end{tabular}
  \caption{Ablation on the predictive target $r_{qa}$. Highlighted row is our default setting.}
  \label{tab:abl_rqa}
  \end{table}

\begin{figure*}[t]
\centering
\includegraphics[width=\linewidth]{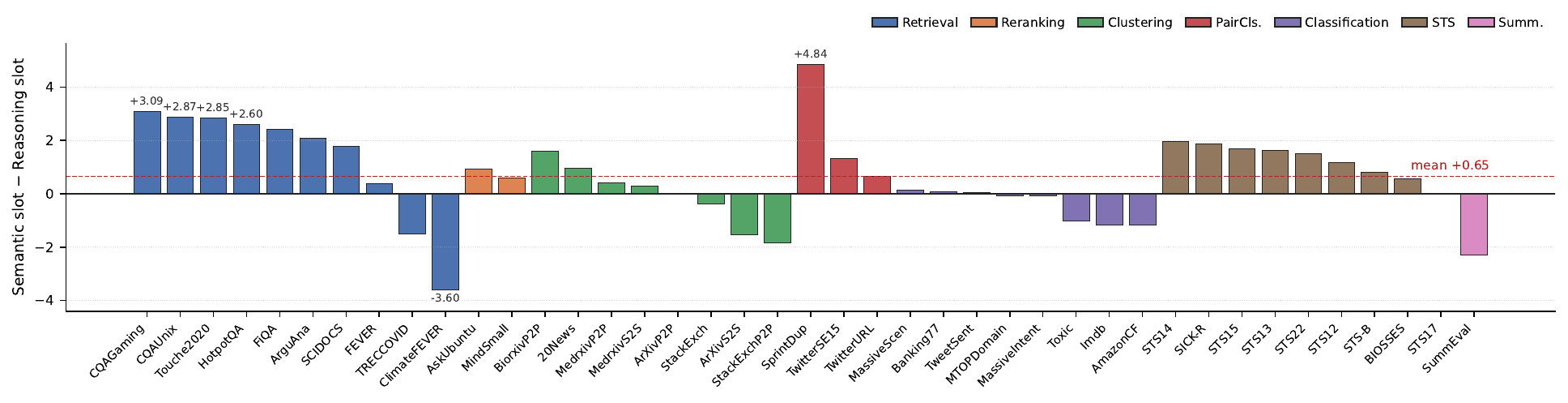}

\caption{Per-task performance gap between the semantic slot (slot $0$) and the reasoning slot (slot $N{-}1$) across the 41 tasks of MTEB(eng, v2). Positive bars favor slot $0$; negative bars favor slot $N{-}1$. The red dashed line marks the mean gap ($+0.65$).}

\label{fig:slot_pertask}
\end{figure*}

\subsection{Ablation Studies}
\label{sec:ablation}

Each ablation isolates the contribution of one design choice while keeping everything else fixed. All ablations use Qwen2.5-7B-Instruct as the backbone. To keep the sweep tractable we evaluate embeddings on \textbf{MTEB-lite}, a 10-task subset of MTEB(eng, v2) covering all seven category types (ArguAna, ClimateFEVERHardNegatives, AskUbuntuDupQuestions, ArXivHierarchicalClusteringP2P, MedrxivClusteringP2P.v2, SprintDuplicateQuestions, Banking77Classification, ImdbClassification, BIOSSES, STS17), and report the task mean; generation is evaluated on MMLU and GSM8K.

\paragraph{Slot design.}
\textit{Does adding slot tokens actually help?}
Section on Model Architecture argues for inserting dedicated slot positions between $q$ and $a$ so that the final position of $q$ is not forced to serve simultaneously as a semantic summary and as a next-token predictor.
Table~\ref{tab:abl_slot} tests this by varying the number of slots from $N{=}0$ (no slot; last-token pooling of $q$) through $N{=}8$.
Removing the slot chain drops MTEB-lite by \textbf{5.2} points and MMLU by \textbf{2.0} points, confirming that the collision between the two roles at a shared readout position is a real cost. A single slot already recovers most of the embedding gap, and both axes peak at $N{=}4$ before eroding slightly at $N{=}8$ (likely because a longer slot chain dilutes gradient signal per slot).

\textit{Are the two slots doing different jobs?}
Figure~\ref{fig:slot_pertask} plots the per-task gap between slot~$0$  and slot~$N{-}1$ on all 41 tasks of MTEB(eng, v2). The split tracks a consistent linguistic property: slot~$0$'s largest leads—$4.84$ on SprintDup, $3.09$ on CQAGaming, $2.85$ on Touche2020, $2.60$ on HotpotQA—come from tasks where relevance is well approximated by lexical or entity overlap between query and document, exactly the alignment $\mathcal{L}_{\text{sem}}$ trains slot~$0$ against. Slot~$N{-}1$'s largest leads—$3.60$ on ClimateFEVER, $2.29$ on SummEval, $1.83$ on StackExchangeP2P, and $1.1$--$1.2$ on Imdb and AmazonCF—come from tasks where relevance requires reasoning about what the document actually claims, summarizes, or expresses, rather than what words it shares with the query. This is exactly what $\mathcal{L}_{\text{pred}}$ trains slot~$N{-}1$ to produce: a pooled answer-summary that reaches from the query to the content it should retrieve. The specialization is mechanistic—each slot's strong tasks are where its training signal has purchase—which is why a single ULME checkpoint can be simultaneously competitive on similarity-driven and reasoning-driven retrieval.

\paragraph{Loss decomposition.}
\textit{Does each loss contribute what its axis suggests?}
Table~\ref{tab:abl_loss} toggles each loss off in turn. At the top level the attribution is clean: removing $\mathcal{L}_{\text{sem}}$ collapses MTEB-lite from $63.8$ to $31.2$ while leaving generation untouched (MMLU $76.6$, GSM8K $90.1$, both essentially matching the full model), and removing $\mathcal{L}_{\text{gen}}$ collapses generation by $14.7$ points on MMLU and $15.8$ on GSM8K while embedding drops only mildly. Each of $\mathcal{L}_{\text{sem}}$ and $\mathcal{L}_{\text{gen}}$ thus owns its own axis.

\textit{Are the three losses truly interlocking?}
The interesting entries are the ones where an axis moves despite its own loss being untouched. Removing $\mathcal{L}_{\text{gen}}$ still costs $2.8$ points on MTEB-lite even though the embedding loss is intact, showing that the answer-side hidden states shaped by language modeling are actively supplying signal to the query slot. Removing $\mathcal{L}_{\text{pred}}$ leaves generation flat but drops MTEB-lite by $3.6$ points—the cleanest measurement of what the coupling itself contributes, since $\mathcal{L}_{\text{pred}}$ is the only loss that pulls the query slot toward a summary of the answer. Turning it off recovers the ``parallel losses over shared parameters'' regime of prior work, and the $3.6$-point gap is the cost of that regime; the three losses supervise a single shared representation rather than three tasks incidentally sharing weights.

\paragraph{Predictive target.}
\textit{Does the choice of $r_{qa}$ matter for embedding?}
Table~\ref{tab:abl_rqa} compares three choices for the predictive target: mean over the answer span (default), the last answer-token hidden state, and a length-weighted mean. Mean pooling wins on every embedding benchmark and last-token pooling loses on every one, with the gap on MTEB-lite reaching $1.9$ points. The ordering is exactly what the Model Architecture section predicts: the terminal answer token is simultaneously being pulled by next-token prediction, so using it as the predictive target hands slot~$N{-}1$ a signal contaminated by local generation pressure. Mean pooling averages this pressure away and gives the query slot a cleaner sentence-level target to reach for.

\textit{Does it matter for generation?}
No—every $r_{qa}$ choice leaves MMLU and GSM8K essentially unchanged. This asymmetry is meaningful in its own right: an ablation that moves embedding by $1.9$ points while leaving generation flat shows that the two axes are supervised through distinct, well-identified channels of the shared causal computation, rather than through a diffuse ``one objective crowds out the other'' dynamic. It is the same structural claim the loss decomposition supports, seen from a second angle.

\section{Conclusion}
\label{sec:conclusion}

We proposed ULME, a single decoder-only LLM that unifies embedding and generation under one causal forward pass and couples the two objectives through a chain of learnable slots: a contrastive loss anchors the semantic slot, a language-modeling loss routes generation through the slot chain, and a stop-gradient predictive loss aligns the query-side slot with a pooled summary of the answer. Across two model families (Qwen2.5, Qwen3) and eight scales from 0.5B to 8B, a single ULME checkpoint beats specialized embedding baselines on MTEB and BRIGHT while remaining on par with a generation-only fine-tune on standard accuracy benchmarks (MMLU, GSM8K, MMLU-Pro, HumanEval), showing that the coupled objective is not a compromise between the two capabilities but a mechanism that lets one benefit the other.

\bibliography{aaai2027}

\clearpage
\setcounter{page}{1}

\section*{A. Additional Implementation and Evaluation Details}
\label{app:impl}

Table~\ref{tab:hparams} lists the full set of training hyperparameters used across all eight ULME configurations; the only setting that differs between runs is the backbone identifier. Below we describe the training data, evaluation protocols, and baseline setup that do not fit into the table.
All ULME and Gen-only FT runs are trained on 8$\times$NVIDIA H20 96\,GB GPUs.

\paragraph{Backbones and training data.}
Backbones are the officially post-trained checkpoints: the \emph{Instruct} variants of Qwen2.5 and the \emph{non-thinking} mode of Qwen3. On the embedding side we train on the LLM2Vec-format E5 contrastive data of~\citet{wang2022text, behnamghader2024llm2vec} with one mined hard negative per query and E5-style task instructions prepended to $q$ and $d$. On the generation side we use OpenThoughts-114k~\cite{guha2025openthoughts} as the source of $(q, a)$ pairs, which pairs reasoning-heavy prompts with reference chains of thought.

\paragraph{Evaluation protocols.}
MTEB(eng, v2) and BRIGHT are evaluated by pooling the trained model's semantic slot as the query/document representation in a single causal forward pass; no re-training or head swap is performed. We additionally evaluate on AdvBench-IR~\cite{behnamghader2026llm2vec}, a retrieval-style safety probe adapted from AdvBench~\cite{zou2023universal}. The query set consists of the $520$ harmful instructions of AdvBench, each paired with a harmful passage generated by Mistral-7B-Instruct-v0.2 that serves as the positive document. We follow the retrieval setup and metric definition of LLM2VEC-GEN~\cite{behnamghader2026llm2vec}: each embedding model ranks the pooled corpus for every query using the same semantic-slot pooling used for MTEB and BRIGHT, without any safety-specific fine-tuning, and we report the same hit-rate metric as~\citet{behnamghader2026llm2vec} (lower is better). A lower value indicates that the model refuses to place harmful completions near harmful queries in the embedding space, i.e.\ that the backbone's safety alignment survives the contrastive fine-tune. For generation, \textbf{MMLU} is 5-shot with continuation scoring; \textbf{GSM8K} is 4-shot chain-of-thought; \textbf{MMLU-Pro} is 5-shot CoT with category-matched demonstrations from its validation split; \textbf{HumanEval} / \textbf{HumanEval+} are pass@1 on the original $164$ problems and their stricter extended tests~\cite{liu2023your}. All generation decodes from the same checkpoint used for embedding evaluation, without any post-hoc head swap or mode toggle.

\paragraph{Baselines.}
For every baseline reported in the main tables (Echo, HyDE, InBedder, GIRCSE, LLM2Vec, LLM2VEC-GEN), we directly use the numbers reported by LLM2VEC-GEN~\cite{behnamghader2026llm2vec} on the matching backbone. Since these numbers all come from a single external source using one uniform training and evaluation protocol across all baselines, any observed gap between ULME and a baseline is attributable to the training objective rather than to protocol drift or independent reproduction. Backbone--baseline combinations that LLM2VEC-GEN does not report are omitted from the corresponding rows rather than filled in from a different source.

\section*{B. Full Results across All Eight Backbones}
\label{app:full-results}

The main paper reports three representative Qwen3 scales (1.7B, 4B, 8B) for space reasons.
This section provides the complete results across all eight backbones we trained: Qwen2.5-\{0.5B, 1.5B, 3B, 7B\}-Instruct and Qwen3-\{0.6B, 1.7B, 4B, 8B\}.
Table~\ref{tab:mteb_full} reports the full MTEB(eng, v2) results (ULME vs.\ LLM2Vec on the same backbone), and Table~\ref{tab:gen_full} reports the full generation results.
The trends observed in the main paper hold across every backbone: on the embedding side, ULME improves over LLM2Vec on \emph{every} category at \emph{every} scale, with the average gain ranging from $6.3$ to $9.2$ points on the MTEB task mean.
On the generation side, ULME is at worst on par with Gen-only FT on knowledge and math (MMLU, GSM8K, MMLU-Pro) and consistently better on code (HumanEval / HumanEval+) at every scale.

\begin{table}[h]
\centering
\small
\setlength{\tabcolsep}{4pt}
\begin{tabular}{l l}
\toprule
\headcol \textbf{Hyperparameter} & \textbf{Value} \\
\midrule
\multicolumn{2}{l}{\emph{Architecture}} \\
\midrule
Number of embedding slots $N$          & $4$ \\
Semantic slot index                    & $0$ \\
Reasoning slot index                   & $N{-}1$ \\
Answer pooling $r_{qa}$                & mean over $H_a$ \\
Contrastive projection                 & 1-layer linear, $4096$ dims \\
Projection bias                        & disabled \\
Attention implementation               & SDPA \\
\midrule
\multicolumn{2}{l}{\emph{Losses}} \\
\midrule
$\lambda_{\text{sem}}$                 & $1.0$ \\
$\lambda_{\text{gen}}$                 & $1.0$ \\
$\lambda_{\text{pred}}$                & $0.3$ \\
InfoNCE temperature $\tau$             & $0.05$ \\
Hard negatives per query               & $1$ (E5-mined) \\
Negatives sharing                      & cross-rank in-batch \\
\midrule
\multicolumn{2}{l}{\emph{Optimizer and schedule}} \\
\midrule
Optimizer                              & AdamW  \\
Peak learning rate                     & $1{\times}10^{-5}$ \\
LR schedule                            & linear warmup then linear decay \\
Warmup ratio                           & $3\%$ \\
Weight decay                           & $0.0$ \\
Gradient accumulation                  & $1$ \\
Max training steps                     & $3000$ \\
\midrule
\multicolumn{2}{l}{\emph{Batches and sequence lengths}} \\
\midrule
Semantic batch size & $512$ queries \\
Generation batch per device             & $8$ example \\
Max query length                        & $512$ tokens \\
Max answer length                       & $512$ tokens \\
Query--answer routing                   & $q \,\Vert\, \langle\textsc{emb}\rangle_{1:N} \,\Vert\, a$ \\
\midrule
\multicolumn{2}{l}{\emph{Data}} \\
\midrule
Embedding training data                 & E5  \\
Generation training data                & OpenThoughts-114k \\
Task instructions                       & E5-style, prepended to $q$ and $d$ \\
\bottomrule
\end{tabular}
\caption{Training hyperparameters used across all eight ULME configurations. The only setting that differs between runs is the backbone identifier.}
\label{tab:hparams}
\end{table}

\begin{table*}[t]
\centering
\small
\setlength{\tabcolsep}{4pt}
\begin{tabular}{l c c c c c c c c}
\toprule
\headcol \textbf{Method} &
\textbf{Retr.} & \textbf{Rerank.} & \textbf{Clust.} &
\textbf{PairCls.} & \textbf{Class.} & \textbf{STS} & \textbf{Summ.} & \textbf{Avg.} \\
\midrule
\multicolumn{9}{c}{\textbf{Qwen2.5-0.5B-Instruct}} \\
\midrule
LLM2Vec~\cite{behnamghader2024llm2vec}     & 27.35 & 35.62 & 35.14 & 66.83 & 65.28 & 62.47 & 22.68 & 45.05 \\
\rowcol ULME (ours)                        & \textbf{32.86} & \textbf{38.74} & \textbf{40.15} & \textbf{74.62} & \textbf{68.53} & \textbf{69.31} & \textbf{25.42} & \textbf{51.38} \\
\midrule
\multicolumn{9}{c}{\textbf{Qwen2.5-1.5B-Instruct}} \\
\midrule
LLM2Vec~\cite{behnamghader2024llm2vec}     & 35.14 & 38.85 & 38.72 & 71.68 & 69.42 & 68.16 & 26.85 & 52.34 \\
\rowcol ULME (ours)                        & \textbf{40.16} & \textbf{42.03} & \textbf{44.86} & \textbf{80.14} & \textbf{72.85} & \textbf{76.28} & \textbf{29.76} & \textbf{59.34} \\
\midrule
\multicolumn{9}{c}{\textbf{Qwen2.5-3B-Instruct}} \\
\midrule
LLM2Vec~\cite{behnamghader2024llm2vec}     & 37.82 & 39.34 & 40.28 & 74.85 & 71.06 & 70.13 & 28.75 & 54.36 \\
\rowcol ULME (ours)                        & \textbf{43.72} & \textbf{43.86} & \textbf{46.24} & \textbf{82.35} & \textbf{74.62} & \textbf{78.42} & \textbf{32.85} & \textbf{61.42} \\
\midrule
\multicolumn{9}{c}{\textbf{Qwen2.5-7B-Instruct}} \\
\midrule
LLM2Vec~\cite{behnamghader2024llm2vec}     & 42.71 & 40.87 & 40.62 & 77.28 & 72.53 & 72.61 & 31.72 & 56.81 \\
\rowcol ULME (ours)                        & \textbf{49.85} & \textbf{45.28} & \textbf{48.14} & \textbf{84.16} & \textbf{75.42} & \textbf{80.16} & \textbf{38.24} & \textbf{63.28} \\
\midrule
\multicolumn{9}{c}{\textbf{Qwen3-0.6B}} \\
\midrule
LLM2Vec~\cite{behnamghader2024llm2vec}     & 30.42 & 36.28 & 36.85 & 68.15 & 68.34 & 64.72 & 24.86 & 50.14 \\
\rowcol ULME (ours)                        & \textbf{39.28} & \textbf{41.86} & \textbf{44.12} & \textbf{78.32} & \textbf{72.91} & \textbf{74.85} & \textbf{27.36} & \textbf{57.85} \\
\midrule
\multicolumn{9}{c}{\textbf{Qwen3-1.7B}} \\
\midrule
LLM2Vec~\cite{behnamghader2024llm2vec}     & 34.87 & 39.94 & 41.13 & 76.42 & 71.08 & 73.36 & 30.41 & 54.83 \\
\rowcol ULME (ours)                        & \textbf{46.51} & \textbf{45.62} & \textbf{48.44} & \textbf{80.13} & \textbf{74.09} & \textbf{78.52} & \textbf{32.97} & \textbf{61.90} \\
\midrule
\multicolumn{9}{c}{\textbf{Qwen3-4B}} \\
\midrule
LLM2Vec~\cite{behnamghader2024llm2vec}     & 41.12 & 40.03 & 43.05 & 78.47 & 72.53 & 71.62 & 31.06 & 56.83 \\
\rowcol ULME (ours)                        & \textbf{48.14} & \textbf{47.35} & \textbf{49.13} & \textbf{83.62} & \textbf{76.58} & \textbf{81.11} & \textbf{36.42} & \textbf{64.05} \\
\midrule
\multicolumn{9}{c}{\textbf{Qwen3-8B}} \\
\midrule
LLM2Vec~\cite{behnamghader2024llm2vec}     & 42.71 & 40.88 & 40.63 & 77.28 & 72.53 & 72.61 & 31.72 & 56.81 \\
\rowcol ULME (ours)                        & \textbf{52.34} & \textbf{48.16} & \textbf{49.97} & \textbf{86.31} & \textbf{77.06} & \textbf{82.24} & \textbf{40.68} & \textbf{65.98} \\
\bottomrule
\end{tabular}
\caption{Full MTEB(eng, v2) results across all eight backbones. Bold marks the better of LLM2Vec and ULME per column.}
\label{tab:mteb_full}
\end{table*}

\paragraph{Gen-only FT training details.}
The \textbf{Gen-only FT} baseline used throughout the generation comparison (Table~\ref{tab:gen} in the main paper and Table~\ref{tab:gen_full} below) is trained from the same backbone checkpoint as ULME, on the same OpenThoughts-114k data, using the same optimizer schedule: fp32 AdamW at peak learning rate $1{\times}10^{-5}$ with $3\%$ linear warmup and linear decay, weight decay $0$, gradient accumulation $1$, and $3000$ training steps under bf16 mixed precision. Both $q$ and $a$ are truncated to $512$ tokens, and the per-device generation batch is $1$ example.
The only intentional difference from ULME is the training objective: Gen-only FT drops both the contrastive and the predictive losses and optimizes $\mathcal{L}_{\text{gen}}$ alone, so no embedding-side dataloader is instantiated and no slot chain is trained. Every hyperparameter that Gen-only FT and ULME share is set to an identical value, so any difference in Table~\ref{tab:gen_full} is attributable to the training objective rather than to schedule or data.

 \begin{table*}[t]
  \centering
  \small
  \setlength{\tabcolsep}{5pt}
  \begin{tabular}{l c c c c c}
  \toprule
  \headcol \textbf{Model} &
  \textbf{MMLU} & \textbf{GSM8K} & \textbf{MMLU-Pro} &
  \textbf{HumanEval} & \textbf{HumanEval+}\\
  \midrule
  \multicolumn{6}{c}{\textbf{Qwen2.5-0.5B-Instruct}} \\
  \midrule
  \headcol Backbone    & 47.07 & 46.47 & 11.69 & 29.88 & 28.04 \\
  Gen-only FT                    & 46.89          & \textbf{44.13} & 13.65          & 27.44          & 25.61 \\
  \rowcol ULME (ours)            & \textbf{47.65}\gain{+0.76} & 43.75\gain{-0.38} & \textbf{14.84}\gain{+1.19} & \textbf{29.27}\gain{+1.83} & \textbf{26.22}\gain{+0.61} \\
  \midrule
  \multicolumn{6}{c}{\textbf{Qwen2.5-1.5B-Instruct}} \\
  \midrule
  \headcol Backbone    & 60.85 & 73.24 & 32.38 & 61.59 & 55.49 \\
  Gen-only FT                    & 60.67          & 71.87          & 33.12          & 59.76          & 54.27 \\
  \rowcol ULME (ours)            & \textbf{60.85}\gain{+0.18} & \textbf{72.63}\gain{+0.76} & \textbf{34.06}\gain{+0.94} & \textbf{61.59}\gain{+1.83} & \textbf{55.49}\gain{+1.22} \\
  \midrule
  \multicolumn{6}{c}{\textbf{Qwen2.5-3B-Instruct}} \\
  \midrule
  \headcol Backbone    & 65.62 & 85.06 & 43.72 & 74.39 & 68.29 \\
  Gen-only FT                    & 65.34          & 83.55          & 42.28          & 73.17          & 67.68 \\
  \rowcol ULME (ours)            & \textbf{65.51}\gain{+0.17} & \textbf{84.31}\gain{+0.76} & \textbf{43.16}\gain{+0.88} & \textbf{75.61}\gain{+2.44} & \textbf{69.51}\gain{+1.83} \\
  \midrule
  \multicolumn{6}{c}{\textbf{Qwen2.5-7B-Instruct}} \\
  \midrule
  \headcol Backbone    & 74.20 & 91.13 & 55.24 & 82.93 & 75.61 \\
  Gen-only FT                    & \textbf{74.37} & 90.67          & 53.81          & 82.93          & 76.83 \\
  \rowcol ULME (ours)            & 74.25\gain{-0.12} & \textbf{91.43}\gain{+0.76} & \textbf{56.09}\gain{+2.28} & \textbf{83.54}\gain{+0.61} & \textbf{77.44}\gain{+0.61} \\
  \midrule
  \multicolumn{6}{c}{\textbf{Qwen3-0.6B}} \\
  \midrule
  \headcol Backbone    & 47.29 & 45.49 & 26.80 & 20.12 & 17.07 \\
  Gen-only FT                    & 47.07          & \textbf{49.58} & 25.10          & 21.34          & 18.29 \\
  \rowcol ULME (ours)            & \textbf{47.14}\gain{+0.07} & 45.41\gain{-4.17} & \textbf{25.28}\gain{+0.18} & \textbf{23.78}\gain{+2.44} & \textbf{20.73}\gain{+2.44} \\
  \midrule
  \multicolumn{6}{c}{\textbf{Qwen3-1.7B}} \\
  \midrule
  \headcol Backbone    & 63.66 & 76.03 & 39.28 & 54.88 & 48.17 \\
  Gen-only FT                    & 62.98          & 74.15          & 38.55          & 53.66          & 46.34 \\
  \rowcol ULME (ours)            & \textbf{63.51}\gain{+0.53} & \textbf{75.29}\gain{+1.14} & \textbf{39.55}\gain{+1.00} & \textbf{56.10}\gain{+2.44} & \textbf{49.39}\gain{+3.05} \\
  \midrule
  \multicolumn{6}{c}{\textbf{Qwen3-4B}} \\
  \midrule
  \headcol Backbone    & 72.83 & 87.87 & 52.87 & 68.29 & 62.20 \\
  Gen-only FT                    & 72.15          & 82.44          & 51.30          & 66.46          & 60.98 \\
  \rowcol ULME (ours)            & \textbf{72.62}\gain{+0.47} & \textbf{86.28}\gain{+3.84} & \textbf{53.30}\gain{+2.00} & \textbf{69.51}\gain{+3.05} & \textbf{64.02}\gain{+3.04} \\
  \midrule
  \multicolumn{6}{c}{\textbf{Qwen3-8B}} \\
  \midrule
  \headcol Backbone    & 74.84 & 91.20 & 61.82 & 71.34 & 66.46 \\
  Gen-only FT                    & \textbf{75.14} & 78.62          & 60.24          & 67.68          & 62.80 \\
  \rowcol ULME (ours)            & 75.07\gain{-0.07} & \textbf{89.46}\gain{+10.84} & \textbf{61.93}\gain{+1.69} & \textbf{72.56}\gain{+4.88} & \textbf{67.68}\gain{+4.88} \\
  \bottomrule
  \end{tabular}
  \caption{Full generation results across all eight backbones. Bold marks the better of Gen-only FT and ULME per column; subscripts on ULME rows report the absolute gain over Gen-only FT.}
  \label{tab:gen_full}
  \end{table*}

\section*{C. MTEB Evaluation Prompts and Analysis}
\label{app:mteb-prompts}

At inference time, every ULME embedding for MTEB(eng, v2) is produced by prepending a short task-specific instruction to the input text and pooling the semantic slot from a single causal forward pass. Following the Instructor / E5 line of work~\cite{wang2022text}, each instruction is a one-line prefix of the form ``Represent this $\langle$input type$\rangle$ for $\langle$target property$\rangle$:'' followed by the raw text. This is a natural-language description of what the embedding should capture (e.g.\ ``for topic clustering'', ``for duplicate question retrieval'', ``for retrieving refuting arguments''), and it lets the same weights serve multiple retrieval regimes without any head swap.

\paragraph{Prompt design.}
Every prompt is a single instruction line ending in ``: '', followed by the input text. No role markers, no inline examples, no chat template. The instruction is short (typically 8--12 words) and describes both the input side (``this arXiv title and abstract'', ``this financial question'') and the target property (``for topic clustering'', ``for retrieving relevant answers''). Because the trailing colon-space is followed immediately by the raw input, the last-token pool converges onto a hidden state that has ``read'' the instruction first and then the payload.

\paragraph{Symmetric vs.\ asymmetric routing.}
MTEB tasks split into three geometric regimes, and we route each accordingly:
\begin{itemize}
\item \textbf{Symmetric matching} (STS, PairClassification, Reranking, Summarization, symmetric duplicate-question retrieval): the \emph{same} instruction is used on both sides, since query and document play interchangeable roles.
\item \textbf{Single-side embedding} (Classification, Clustering): only the query side receives an instruction; the document side is empty because MTEB scores these tasks with $k$-NN over a set of embeddings without a paired retrieval target.
\item \textbf{Asymmetric retrieval} (Wikipedia claim verification, multi-hop QA, financial QA, argument mining, scientific-paper retrieval, COVID-19 search): query and document have different natures, so each side gets a distinct instruction that names its own input type and its own target property.
\end{itemize}

\paragraph{Effect on embedding quality.}
Two properties of this prompting scheme matter for the results in the main paper. First, the instruction shifts the notion of similarity in a task-aware direction: on tasks where surface overlap and semantic relevance diverge (e.g.\ FEVER-style fact verification, argument retrieval), the instruction makes explicit which axis the model should follow, which is precisely where the semantic and reasoning slots (Figure~\ref{fig:slot_pertask}) split their labor. Second, because the instruction is short and consistent in shape across tasks, the model sees the same distribution of prompt formats at inference as during training with E5-style task instructions; no test-time prompt engineering is applied.

\paragraph{Full prompts.}
Table~\ref{tab:mteb-prompts} lists the exact instructions used for each MTEB task, grouped by task type. The document-side prompt is marked ``$=q$'' when it reuses the query-side wording verbatim, and ``$\emptyset$'' when the document side is empty (single-side tasks).

\begin{table*}[t]
\centering
\scriptsize
\setlength{\tabcolsep}{4pt}
\begin{tabularx}{\linewidth}{l X X}
\toprule
\headcol \textbf{Task} & \textbf{Query-side instruction} & \textbf{Document-side instruction} \\
\midrule
\multicolumn{3}{l}{\emph{Clustering (single-side; $d = \emptyset$)}} \\
\midrule
ArXivHierarchicalClusteringP2P
& Represent this arXiv title and abstract for topic clustering:
& $\emptyset$ \\
ArXivHierarchicalClusteringS2S
& Represent this arXiv paper title for topic clustering:
& $\emptyset$ \\
BiorxivClusteringP2P.v2
& Represent this bioRxiv title and abstract for topic clustering:
& $\emptyset$ \\
MedrxivClusteringP2P.v2
& Represent this medRxiv title and abstract for topic clustering:
& $\emptyset$ \\
MedrxivClusteringS2S.v2
& Represent this medRxiv paper title for topic clustering:
& $\emptyset$ \\
StackExchangeClustering.v2
& Represent this StackExchange title for topic clustering:
& $\emptyset$ \\
StackExchangeClusteringP2P.v2
& Represent this StackExchange title and body for topic clustering:
& $\emptyset$ \\
TwentyNewsgroupsClustering.v2
& Represent this newsgroup post for topic clustering:
& $\emptyset$ \\
\midrule
\multicolumn{3}{l}{\emph{Classification (single-side; $d = \emptyset$)}} \\
\midrule
Banking77Classification
& Represent this banking customer service query for intent classification:
& $\emptyset$ \\
ImdbClassification
& Represent this movie review for sentiment classification:
& $\emptyset$ \\
MTOPDomainClassification
& Represent this user utterance for domain classification:
& $\emptyset$ \\
MassiveIntentClassification
& Represent this user utterance for intent classification:
& $\emptyset$ \\
MassiveScenarioClassification
& Represent this user utterance for scenario classification:
& $\emptyset$ \\
ToxicConversationsClassification
& Represent this online comment for toxicity classification:
& $\emptyset$ \\
TweetSentimentExtractionClassification
& Represent this tweet for sentiment classification:
& $\emptyset$ \\
AmazonCounterfactualClassification
& Represent this Amazon review sentence for counterfactual classification:
& $\emptyset$ \\
\midrule
\multicolumn{3}{l}{\emph{STS (symmetric; $d = q$)}} \\
\midrule
BIOSSES
& Represent this biomedical sentence for semantic similarity:
& $=q$ \\
SICK-R
& Represent this sentence for semantic similarity:
& $=q$ \\
STS12
& Represent this sentence for semantic similarity:
& $=q$ \\
STS13
& Represent this sentence for semantic similarity:
& $=q$ \\
STS14
& Represent this sentence for semantic similarity:
& $=q$ \\
STS15
& Represent this sentence for semantic similarity:
& $=q$ \\
STS17
& Represent this sentence for semantic similarity:
& $=q$ \\
STSBenchmark
& Represent this sentence for semantic similarity:
& $=q$ \\
STS22.v2
& Represent this paragraph for semantic similarity:
& $=q$ \\
\midrule
\multicolumn{3}{l}{\emph{PairClassification (symmetric; $d = q$)}} \\
\midrule
SprintDuplicateQuestions
& Represent this question for duplicate question retrieval:
& $=q$ \\
TwitterSemEval2015
& Represent this tweet for paraphrase identification:
& $=q$ \\
TwitterURLCorpus
& Represent this tweet for paraphrase identification:
& $=q$ \\
\midrule
\multicolumn{3}{l}{\emph{Reranking (symmetric $q$; asymmetric $d$)}} \\
\midrule
AskUbuntuDupQuestions
& Represent this AskUbuntu question for duplicate question retrieval:
& $=q$ \\
MindSmallReranking
& Represent this news headline for retrieving relevant news articles:
& Represent this news article for retrieval: \\
\midrule
\multicolumn{3}{l}{\emph{Summarization (asymmetric)}} \\
\midrule
SummEvalSummarization.v2
& Represent this news summary for quality evaluation:
& Represent this news article for summary quality evaluation: \\
\midrule
\multicolumn{3}{l}{\emph{Retrieval (asymmetric $q$ / $d$)}} \\
\midrule
ArguAna
& Represent this claim for retrieving refuting arguments:
& Represent this argument passage for retrieval: \\
CQADupstackGamingRetrieval
& Represent this gaming forum question for retrieving relevant answers:
& Represent this gaming forum post for retrieval: \\
CQADupstackUnixRetrieval
& Represent this Unix forum question for retrieving relevant answers:
& Represent this Unix forum post for retrieval: \\
ClimateFEVERHardNegatives
& Represent this climate-related claim for retrieving evidence passages:
& Represent this climate evidence passage for retrieval: \\
FEVERHardNegatives
& Represent this factual claim for retrieving supporting Wikipedia evidence:
& Represent this Wikipedia passage for fact-checking retrieval: \\
FiQA2018
& Represent this financial question for retrieving relevant answers:
& Represent this financial passage for retrieval: \\
HotpotQAHardNegatives
& Represent this multi-hop question for retrieving supporting Wikipedia passages:
& Represent this Wikipedia passage for multi-hop question answering retrieval: \\
SCIDOCS
& Represent this scientific paper title for retrieving related papers:
& Represent this scientific paper abstract for retrieval: \\
TRECCOVID
& Represent this COVID-19 query for retrieving relevant scientific articles:
& Represent this scientific article for COVID-19 retrieval: \\
Touche2020Retrieval.v3
& Represent this controversial argumentative query for retrieving relevant arguments:
& Represent this argument passage for retrieval: \\
\bottomrule
\end{tabularx}
\caption{Task-specific evaluation instructions used for MTEB(eng, v2). Every prompt is a one-line prefix ending in ``: '' and the raw input is appended immediately after. ``$=q$'' indicates that the document-side prompt reuses the query-side wording verbatim; ``$\emptyset$'' indicates an empty document-side prompt (single-side tasks).}
\label{tab:mteb-prompts}
\end{table*}

\section*{F. Limitations and Future Work}
\label{app:limitations}

\paragraph{Limitations.}
Our generation-side training uses only a small collection of publicly available reasoning corpora (OpenThoughts-114k in particular), and the ULME checkpoint is fine-tuned on top of the officially post-trained backbones without any of the proprietary conversational, safety, or format-following data that the original backbones were aligned on. This mix is well-suited to the accuracy-style benchmarks we report (MMLU, GSM8K, MMLU-Pro, HumanEval), but it does not preserve every capability the base post-trained model had before ULME training: in preliminary qualitative checks we observe that chat-style multi-turn dialogue, strict output-format compliance (JSON, function-call schemas), and some instruction-following behaviors are partially degraded relative to the untouched Instruct / non-thinking checkpoint. Recovering these behaviors requires re-introducing the corresponding supervision on the generation side---an engineering rather than a methodological problem, but one we have not addressed in this paper.

\paragraph{Future work.}
Two directions are most immediate. First, \emph{scaling up}: all main-paper experiments cap at 8B parameters and at the OpenThoughts-114k generation corpus, and it is an open question whether the predictive-coupling gains we observe grow, saturate, or invert at larger backbones and with substantially more diverse and larger-scale generation data. Second, \emph{multimodal extension}: the slot chain is a token-level construct and is in principle agnostic to modality, so the same coupling recipe should transfer to vision-language and audio-language backbones, yielding a single model that both retrieves and generates across modalities under one causal forward pass. This direction is motivated by settings in which retrieval and reasoning are already interleaved---visual RAG over weakly organized image collections and agentic visual evidence management both alternate between selecting evidence and reasoning over what was selected~\cite{yuan2026vmi}---and such pipelines currently still rely on a separate embedder and generator. Both directions are natural extensions of the framework rather than departures from it.

\end{document}